%% file: main.tex
\documentclass{article}

\usepackage[preprint]{neurips_2026}

\usepackage[utf8]{inputenc}
\usepackage[T1]{fontenc}
\usepackage{hyperref}
\usepackage{url}
\usepackage{booktabs}
\usepackage{graphicx}
\usepackage{wrapfig}
\usepackage{multirow}
\usepackage{amsmath}
\usepackage{amsfonts}
\usepackage{nicefrac}
\usepackage{microtype}
\usepackage{xcolor}
\usepackage{xspace}
\input{common_commands}

\usepackage{adjustbox}
\usepackage{tcolorbox}
\usepackage{listings}

\lstdefinestyle{redprompt}{
  breaklines=true,
  columns=fullflexible,
  basicstyle=\ttfamily\footnotesize\color{black}
}
\usepackage{capt-of}
\usepackage{tcolorbox}

\newtcolorbox{reasoningbox}[1]{
  colback=white,
  colframe=black!20,
  boxrule=0.4pt,
  arc=1pt,
  left=3pt,
  right=3pt,
  top=2.5pt,
  bottom=2.5pt,
  before upper=\raggedright,
  title=#1,
  fonttitle=\bfseries,
  colbacktitle=black!5,
  coltitle=black
}

\title{Rethinking Post-Training Recipes\\ for Multimodal Time-Series Forecasting}
\author{
Haoxin Liu\textsuperscript{\textdagger,\S}\thanks{Correspondence to: \texttt{hliu763@gatech.edu}. This work was done while the author was a Student Researcher at Google Research.},
Yichen Zhou\textsuperscript{\S},
Rajat Sen\textsuperscript{\S},
B. Aditya Prakash\textsuperscript{\textdagger},
Abhimanyu Das\textsuperscript{\S} \\
\vspace{0.25em}
\parbox{\textwidth}{\centering
\makebox[0.28\textwidth][c]{\textsuperscript{\textdagger}Georgia Institute of Technology}
\hspace{0.01\textwidth}
\makebox[0.36\textwidth][c]{\textsuperscript{\S}Google Research}
}
}

\begin{document}

\maketitle

\begin{abstract}

Time-Series Foundation Models (TSFMs) excel at zero-shot unimodal forecasting using numerical data, but unlike LLMs they cannot consume multimodal, non-numerical context that often shape real-world trajectories. In this work, we bridge this gap and argue for a multimodal time-series forecasting approach that post-trains LLMs to act as context-guided revisors over strong numerical TSFM priors. We introduce \method, a post-training recipe combining Supervised Fine-Tuning (SFT) and Reinforcement Learning with Verifiable Rewards (RLVR), along with a methodology to generate automated reasoning traces for forecast revisions. \method teaches an LLM  to generate context-conditioned forecast interventions---decisions to revise, preserve, or ignore the TSFM prior based on the multimodal context. We evaluate this approach on the TimesX multimodal forecasting benchmark using a Gemma-3-4B LLM and TimesFM-2.5 TSFM, and show that it significantly outperforms standalone TSFMs, LLM-only baselines, and existing multimodal forecasting approaches.

\end{abstract}

\input{Sections/01_introduction}

\input{Sections/02_setup_and_metrics}
\input{Sections/03-New-Preliminary}
\input{Sections/03_diagnostic_study}

\input{Sections/05a_main_results}

\input{Sections/06_further_analysis}

\bibliographystyle{abbrvnat}
\bibliography{citations}

\appendix

\input{Sections/A_recipe_implementation_details}
\input{Sections/05_main_results}

\end{document}

%% file: common_commands.tex
\newcommand{\method}{\textsc{PostTime}\xspace}

\newcommand{\impratio}{\textsc{ImpRatio}\xspace}
\newcommand{\timesfm}{\textsc{TimesFM}\xspace}
\newcommand{\tsfm}{TSFM\xspace}
\newcommand{\trace}{\tau}
\newcommand{\timex}{\textsc{TimesX}\xspace}
\newcommand{\tracegen}{trace generator}

%% file: Sections/01_introduction.tex
\section{Introduction}
\label{sec:introduction}

Time series forecasting (TSF)~\citep{liu2024lingkai} is a foundational problem across diverse domains, from climate modeling and healthcare to finance and supply chain management. Historically reliant on domain-specific statistical models or task-specific neural architectures, the field has recently undergone a paradigm shift with the advent of Time-Series Foundation Models (TSFMs). TSFMs such as Moirai~\citep{woo2024unified}, TimesFM~\citep{das2023decoder} and Chronos~\citep{ansari2025chronos} are pre-trained on vast, diverse corpora of temporal data, allowing them to capture complex temporal dynamics and patterns purely from historical numerical inputs and achieve state-of-the-art, zero-shot performance on  unimodal forecasting benchmarks.

Despite the remarkable success of TSFMs, real-world forecasting is rarely a strictly unimodal endeavor. Future trajectories are frequently shaped by non-numerical, exogenous factors---such as breaking news events, sudden policy changes, shifting metadata, or domain-specific textual descriptions. TSFMs are structurally constrained by their unimodal inputs and cannot directly consume this multimodal non-numeric context. This  highlights an urgent need for multimodal forecasting approaches capable of seamlessly fusing raw numerical histories with rich semantic context.

At the same time, modern Large Language Models (LLMs) have demonstrated an unprecedented ability to understand and process complex multimodal data and execute sophisticated reasoning paths. This makes them  natural candidates for  ingesting and interpreting the unstructured semantic context in multimodal forecasting. However LLMs have been found to be unsuited for raw numerical predictive tasks, and significantly underperform TSFMs as standalone numerical forecasters~\citep{tan2024language,merrill2024language}.

This mismatch highlights the following question: how can we systematically unify the  numerical performance of TSFMs with the multimodal reasoning capabilities of modern LLMs for robust multimodal forecasting? In recent years, there have been several approaches to building multimodal forecasting models  within the design space spanned by TSFMs and LLMs (see \cite{liu2025can} for a comprehensive survey), ranging from converting the temporal modality into other modalities, fusing temporal and non-temporal modalities at the input embedding stage, or at the output of modality-specific models, and pretraining from scratch on a multimodal time series corpus \citep{wu2025aurora}.
In this paper, we focus on a different paradigm: post-training off-the-shelf LLMs in the presence of TSFM priors. This paradigm requires us to define precisely how and what an LLM should learn during post-training, when provided access to a strong numerical TSFM prior. We evaluate and propose various design decisions for this post-training recipe, and showcase how some standard LLM post-training design choices are not as appropriate for multimodal forecasting.

Our methodology (that is supported by various ablation studies) starts with the thesis that treating the LLM as a standalone numerical forecaster on multimodal inputs is suboptimal - the LLM is better positioned as a context-guided reviser over the TSFM prior. However, we observe that simply using an out-of-the-box LLM to perform raw revision of a TSFM prior does not yield good performance. Post-training the LLM in the presence of the TSFM prior, using Supervised Fine Tuning (SFT) or Reinforcement Learning (RL), is critical to unlocking its context-guided revision capabilities. This then opens up several design questions  around this post training methodology: (1) how can we scalably generate good quality SFT traces, (2) how  do we ensure our post-training data covers the full spectrum of the revision policy including when to revise and how to revise the TSFM prior, and (3) how can we construct suitable reward functions for RL to further reinforce successful revision policies and improve the final multimodal forecasting accuracy.

In this work, we address these questions and propose a generic post-training recipe for multimodal forecasting using context-guided revision called \method. This  recipe involves a supervised-fine-tuning stage (SFT) followed by a reinforcement-learning-with-verifiable-rewards stage (RLVR). The post-training recipe does not just train a base LLM to perform direct imitation of the ground truth forecasts, but instead encourages the base LLM to produce effective forecast interventions: context-conditioned decisions that revise, or preserve the strong numerical prior depending on the relative strength of the prior versus the non-numeric context.

We design and evaluate  a specific instantiation of \method involving Gemma-3-4B as the LLM to be post-trained, TimesFM-2.5 as the TSFM prior provider. This is coupled with an automated approach (no human-in-the-loop) for generating reasoning traces using a frontier LLM (Gemini-3.1-Flash-Lite) on multimodal examples from a refresh of the TimesX dataset\citep{liu2026rethinking}.

Our paper evaluates this post-training recipe with a comprehensive set of experiments on the refreshed TimesX benchmark that includes both ablations on the design choices mentioned earlier, and multimodal forecasting accuracy comparisons against a spectrum of forecasting approaches (including strong supervised models, {\tsfm}s, LLM-only models, out-of-the-box LLM + TSFM revision and ensembling methods, recent time-series reasoning models and other multimodal forecasting models.

The dataset, code, and models will be released upon acceptance of the paper.

%% file: Sections/03-New-Preliminary.tex
\section{Preliminaries}
\label{sec:prelim}
\paragraph{Problem Formulation.}\label{sec:setup}We focus on multimodal forecasting of univariate time series. Let $x_{1:T}$ denote the historical time series where $T$ is the length of the history, $c$ the text context (without leakage) available before the forecast horizon, and $y_{1:H}$ the future horizon where $H$ is the length of the horizon. The goal is to produce an accurate forecast $\hat{y}_{1:H}$ using information from both modalities.

The most direct approach is to post-train an LLM policy $\pi_{\theta}$ that takes $x_{1:T}$ and $c$ as inputs and generates the forecast end-to-end,
$$
(\trace, \hat{y}_{1:H}) = \pi_{\theta}(x_{1:T}, c),
$$
where $\trace$ is a reasoning trace (e.g. thoughts for reasoning models).

However, a potentially better strategy could be to explicitly leverage a strong numerical prior from a base forecaster, for instance, a TSFM. The base forecaster produces a forecast prior (a base forecast) using only the numerical history,
$$
\tilde{y}_{1:H} = f_{\text{TSFM}}(x_{1:T}),
$$
and the LLM policy then revises this base forecast using both $x_{1:T}$ and $c$,
$$
(\trace, \hat{y}_{1:H}) = \pi_{\theta}(x_{1:T}, c, \tilde{y}_{1:H})
$$
where $\trace$ is a reasoning trace and $\hat{y}_{1:H}$ is the final forecast. Intuitively, $\tilde{y}_{1:H}$ anchors the prediction to a specialized numerical model, while the LLM focuses on incorporating textual information and making calibrated adjustments. Note that we will drop the temporal subscripts when clear from context and with some abuse of notation use a subscript index like $i$ to denote the $i$-th post-training example.

We will show in Section~\ref{subsec:rethinking_llm_role} that the second strategy is empirically better. Implementation wise, for both strategies the LLM is prompted to output some reasoning trace before a final forecast. See Appendix~\ref{app:prompt_output_format} for the templates.

\paragraph{Dataset and Setting.}
We use a refreshed version of \timex~\citep{liu2026rethinking} built with its public dataset agent. This refresh aligns numerical time-series values with rich textual context, including metadata, calendar signals, covariates, and time-stamped events, covering 99 variables from 2022 to 2025. Following \timex, we use a forecast history length $T = 96$ and a horizon length of $H=12$. We choose a rolling window shift of 4 for weekly data and 12 for daily data to obtain to obtain equal numbers of weekly and daily examples.

To prevent LLM knowledge leakage and future information leakage, we split the data by Jan.~30, 2025, a common knowledge-cutoff date used for training recent LLMs. Our post-training uses samples whose prediction windows end before this cutoff, while evaluation uses samples whose prediction windows start after this cutoff. We further randomly select 88 variables for training and in-domain (ID) evaluation, and hold out the remaining 11 variables for out-of-domain (OOD) evaluation. Appendix~\ref{app:timex_data_demo} provides a concrete data example, and Appendix~\ref{app:recipe_implementation_details} lists all variables in each split.

We use the standard MAE (mean absolute error) and MSE (mean squared error) metrics to measure forecast accuracy. For making fair comparisons across and aggregations over different variables we also normalize them, when applicable, by the same metrics of the seasonal naive forecast which repeats the last period in the history.
\begin{align*}
\text{nMAE}(\hat{y}, y | x) = \frac{\sum_{t=1}^H \left|y_t - \hat{y}_t\right|}{\sum_{t=1}^H \left|y_t - \hat{y}^{\text{seasonal naive}}_t\right|},
~~ \text{nMSE}(\hat{y}, y | x) = \frac{\sum_{t=1}^H \left(y_t - \hat{y}_t\right)^2}{\sum_{t=1}^H \left(y_t - \hat{y}^{\text{seasonal naive}}_t\right)^2}.
\end{align*}

\paragraph{LLM Post-Training.} In this work, LLM post-training refers to updating the weights of a pretrained LLM to strengthen task-specific capabilities. Early LLM post-training recipes mainly rely on SFT, where the model imitates curated prompt-response pairs~\citep{ouyang2022training, chung2024scaling, Wei2021}. Recent recipes start to use Chain-of-Thought (CoT) SFT as a warm-up stage, training the model to produce useful intermediate reasoning before the final answer~\citep{wei2022chain, zelikman2022star}, followed by RL, often with GRPO-style optimization and rule-based rewards, to push the model beyond imitation~\citep{shao2024deepseekmath,guo2025deepseek}.

Under this general post-training paradigm there are design choices further made for solving domain specific tasks such as mathematics, coding, and logic. In this work we would like to rethink these practices in the context of multimodal forecasting and come up with a streamlined post-training recipe for the same.

%% file: Sections/03_diagnostic_study.tex
\section{\textcolor{black}{Systemic Rethinking of Post-Training Recipes}}
\label{sec:diagnostic_study}

This section systemically rethinks post-training recipes for multimodal TSF. We organize the design space into three stages: the role assigned to the LLM, the construction of CoT-SFT supervision, and the RL objective used to refine the resulting policy. For each stage, we contrast current practice with our design point and then provide empirical evidence that motivates the final \method recipe.

\subsection{\textcolor{black}{The Role of LLMs: Reviser is a Better Starting Point}}
\label{subsec:rethinking_llm_role}

\noindent\textbf{\textcolor{black}{Current practice.}}
The first post-training decision is the role assigned to the LLM. The most direct solution is to equip the LLM to generate the forecast conditioned on the time-series history and the text context. This is the paradigm taken by most prior works that post-train an LLM for multimodal forecasting or reasoning~\citep{guan2025timeomni,kong2025timemqa,liu2025timer1}. It forces the LLM to solve numerical extrapolation, context use, and strict forecast-format compliance all at once.

\noindent\textbf{\textcolor{black}{Our point and design.}}
We argue that the LLM is better suited as a reviser of a base forecast generated by a \tsfm from the time-series history. Given history $x$, text context $c$, and an initial forecast $\tilde y$, the LLM decides whether the context warrants a structural revision of the numerical prior. This design simplifies the LLM's reasoning task while retaining the strong numerical prior of capable \tsfm models such as Chronos-2~\citep{ansari2025chronos} and \timesfm-2.5~\citep{das2023decoder}.

\noindent\textbf{\textcolor{black}{Empirical evidence.}}
To establish that revising is a better starting point, we evaluate several powerful LLMs under two zero-shot settings: direct forecasting, $\hat{y}=\pi_{\theta}(x,c)$, and revising, $\hat{y}=\pi_{\theta}(x,c,\tilde{y})$. The results are summarized in Table~\ref{tab:direct_vs_revising}.

\begin{table}[t]
\centering
\caption{Direct forecasting versus revising a \timesfm-2.5 prior on ID evaluation. Revising improves nMAE in 6/8 LLMs and nMSE in 7/8 LLMs. Underlines mark the better value within each LLM pair.}
\label{tab:direct_vs_revising}
\resizebox{\linewidth}{!}{
\begin{tabular}{lrrrr}
\toprule
 & Direct nMAE $\downarrow$ & Direct nMSE $\downarrow$ & Revising nMAE $\downarrow$ & Revising nMSE $\downarrow$ \\
\midrule
\timesfm-2.5 & 0.788 & 0.729 & -- & -- \\
\midrule
GPT-5 & 0.843 & 1.299 & \underline{0.783} & \underline{0.725} \\
Gemini-3.1-Pro & \underline{0.917} & \underline{8.890} & 0.949 & 9.375 \\
Gemini-3.1-Flash-Lite & \underline{0.812} & 3.101 & 0.820 & \underline{2.966} \\
Gemini-3-Flash & 1.022 & 10.014 & \underline{0.950} & \underline{7.349} \\
Gemini-2.5-Flash & 1.261 & 13.445 & \underline{1.151} & \underline{6.716} \\
Qwen-3-VL-32B & 0.947 & 1.122 & \underline{0.915} & \underline{1.026} \\
GPT-OSS-120B & 0.930 & 2.762 & \underline{0.794} & \underline{0.787} \\
Gemma-3-4B & 1.486 & 11.434 & \underline{0.877} & \underline{0.878} \\
\bottomrule
\end{tabular}
}
\end{table}

Revising improves nMAE for 6 of 8 LLMs and nMSE for 7 of 8 LLMs, with both metrics improving for 6 of 8 LLMs. The nMSE gains are especially diagnostic because nMSE is sensitive to large numerical failures: GPT-OSS-120B drops from 2.762 to 0.787, Gemma-3-4B from 11.434 to 0.878, and Gemini-2.5-Flash from 13.445 to 6.716. Revising also improves the valid-window rate from 0.77 for direct forecasting to 0.96 for revising, where the LLM is said to return a valid window when it returns at least as many forecasted points as in the horizon. Thus, a TSFM prior does not only improve numerical stability; it also gives the LLM a more structured output contract for post-training.

Zero-shot revising, however, is not enough. Even the strongest commercial LLMs do not make zero-shot revising a reliable solution: GPT-5 revising slightly improves over \timesfm-2.5 on both ID metrics, but most revisers still trail the \timesfm-2.5 prior. Zero-shot revising exposes the right role for the LLM, but it does not reliably internalize the text conditioned policy for when to revise, how to revise, and when not to revise. In order to really solve the problem we therefore need to post-train an LLM for the revising role and the subsequent sections narrow down on the best practices for that very task.

\input{Sections/03a_rethinking_cot_sft_data}

\subsection{\textcolor{black}{The RL Stage: Reward Improvement Instead of Accuracy}}
\label{subsec:trap_relative_reward}

\noindent\textbf{\textcolor{black}{Current practice.}}
The RL stage is commonly framed as verifiable optimization of final forecast accuracy. Given ground-truth labels, a natural reward maps the model's absolute forecast error to a scalar score. For example, current practice uses Eq.~\ref{eq:expmae_reward}, an exponentially decayed MAE (denoted as ExpMAE) as the verifiable reward.

\begin{equation}
R(\hat{y}, y)
= \exp\!\left(-\frac{\mathrm{MAE}(\hat{y}, y)}{\gamma}\right),
\quad \gamma=10.
\label{eq:expmae_reward}
\end{equation}

However, such a reward can provide weak preference signals under common RL setups such as group-normalized GRPO. With multiple completions sampled for the same prompt, the effective advantage is determined by the reward differences within that prompt group,
\begin{equation}
A_i = \frac{R_i-\mu_{\mathrm{group}}}{\sigma_{\mathrm{group}}+\epsilon_{\mathrm{std}}}.
\label{eq:grpo_group_advantage}
\end{equation}
Thus a directionally correct reward can still lack distinguishing power if it assigns nearly identical scores to multiple completions. Eq.~\ref{eq:expmae_reward} is not scale-free, so the exponential decay can numerically collapse several candidates when they all have high MAEs regardless of their actual forecast quality.

\noindent\textbf{\textcolor{black}{Our point and RL design.}}
For a model revising over a strong \tsfm prior, an objective based solely on final forecast accuracy is incomplete. The reviser is not forecasting from scratch; it is deciding whether the context justifies changing the base forecast. The reward should therefore measure how much a revision improves over the base forecast.

For a generated forecast $\hat y$, we define an improvement ratio against the base forecast as
\begin{equation}
  \mathrm{ImpRatio} = 1-\frac{\mathrm{MAE}(\hat y,y)}
       {\mathrm{MAE}(\tilde{y},y)+\epsilon}, \quad \epsilon > 0,
  \label{eq:impratio_definition}
\end{equation}
and an $\mathrm{ImpRatio}$ based reward as
\begin{equation}
  R(\hat{y}, y) =
  \mathrm{clip}_{[0,1]}
  \left(0.5 + 0.5\times\mathrm{ImpRatio}\right).
  \label{eq:impratio_reward}
\end{equation}
This reward is centered at $0.5$ when the reviser ties with the base forecast. It also linearly separates the final forecasts once they are better than the base forecast, which ensures better resolution for the RL algorithm.

\noindent\textbf{\textcolor{black}{Empirical evidence.}} Table~\ref{tab:rl_reward_signal_reliability} demonstrates ImpRatio is a better reward. The ExpMAE reward has mean group reward standard deviation of only 0.0171, with 27.97\% of prompt groups having near zero variance. In contrast, ImpRatio increases the mean reward standard deviation to 0.0719, and reduces the near-zero-variance group fraction to 4.66\%. ExpMAE also has 16 instances where at least half of the prompt groups provide almost no preference signal, while ImpRatio has none.

\begin{table}[ht]
\centering
\caption{Distinguishing power of ExpMAE and ImpRatio during GRPO training. ImpRatio preserves stronger same-prompt preference signals. Reward STD is computed across completions within the same prompt group. Mean zero-STD frac is the fraction of groups with near-zero reward variance. Collapse steps count logging steps where at least half of prompt groups have near-zero variance.}
\label{tab:rl_reward_signal_reliability}
\begin{tabular}{@{}lccrrr@{}}
\toprule
Reward & Scaling & Revision Driven & Mean STD $\uparrow$ & Zero-STD frac $\downarrow$ & Collapses $\downarrow$ \\
\midrule
ExpMAE & fixed, global & No & 0.0171 & 0.2797 & 16 \\
ImpRatio  & instance base & Yes & \textbf{0.0719} & \textbf{0.0466} & \textbf{0} \\
\bottomrule
\end{tabular}
\end{table}

%% file: Sections/03a_rethinking_cot_sft_data.tex
\color{black}
\subsection{The CoT-SFT Stage: Include Diverse Reasoning Traces and Preserve Hard Cases}
\label{subsec:rethinking_cot_sft_data}
\label{subsec:rethinking_data_sampling}
\label{subsec:rethinking_sft_sample_construction}

In order to kick start a better instruction following behavior for the forecasting revision task, we need to create high quality traces for the SFT phase of post-training. CoT-SFT data construction for multimodal forecasting involves two key decisions: which reasoning traces should be used as imitation targets and which source samples should be kept.

\subsubsection{Reasoning-Trace Annotations: Include Diverse Reasoning Traces}

\paragraph{Current practice.}
The common recipe treats CoT-SFT as imitation of a strictly verifiable reasoning path, which is suitable for domains such as coding, mathematics, and logic where an answer is either correct or wrong. Current recipes for multimodal TSF follow this idea. Time-R1 uses the observed horizon of the prediction window as an input hint for the trace generator~\citep{liu2025timer1}, while Time-Omni~\citep{guan2025timeomni} requires human-in-the-loop traces.

\paragraph{Our point.}

However for multimodal TSF the observed horizon is usually a noisy realization of some ground truth, and one can only hope to reduce this uncertainty in the forecast by conditioning on the context - it is hard to imitate a trace with a retrospective view. Moreover, a single reasoning trace conflicts with the open-ended nature of the future. Therefore, we hide the observed horizon from the trace generator and keep multiple ``valid'' reasoning traces, where validity is defined by whether the final revised forecast improves over the initial forecast.

\begin{wraptable}{r}{0.43\linewidth}
\vspace{-2em}
\centering
\small
\setlength{\tabcolsep}{2pt}
\caption{Imitating the observed horizon hurts performance.}
\label{tab:cot_sft_gt_verifier}
\resizebox{\linewidth}{!}{
\begin{tabular}{@{}lrr@{}}
\toprule
CoT-SFT target & nMAE & nMSE \\
\midrule
Direct observed horizon fitting & 0.881 & 1.266 \\
Observed horizon hinted traces & 0.789 & 0.748 \\
Forecast-time traces & 0.781 & 0.716 \\
\midrule
\timesfm-2.5 & 0.788 & 0.729 \\
\bottomrule
\vspace{-4em}
\end{tabular}
}

\end{wraptable}
\noindent\textbf{\textcolor{black}{Empirical evidence.}} Table~\ref{tab:cot_sft_gt_verifier} supports using the observed horizon as a verifier rather than an imitation target. Direct fitting the observed horizon is much worse than the \timesfm-2.5 prior, and observed-horizon-guided hindsight still trails the prior in nMSE. In contrast, forecast-time traces improve over the prior on both nMAE and nMSE.

\subsubsection{Data Sampling: Preserve Hard Cases}

\paragraph{Current practice.}
Current data sampling practice often keeps only ``easy'' samples. For example, Time-Omni~\cite{guan2025timeomni} selects samples with a relatively low mean absolute error.

\paragraph{Our point.}
In multimodal TSF, hard cases are not merely noisy labels. They often reflect ambiguous or misleading context, where the best action may even be to keep the \tsfm prior. A robust reviser must learn when to revise cautiously, and when to preserve the unimodal forecast.

To expose this structure, we sample five independent forecast-time revisions for each source sample and count how many of them improve over the \timesfm-2.5 initial forecast. A 5/5 bucket indicates an easy revision case with clear contextual signal; 1/5--3/5 indicates a hard or ambiguous case where only some reasoning trajectories find a useful intervention; 0/5 indicates a non-revisable case where fallback is the appropriate supervised action.

\begin{wraptable}{r}{0.63\linewidth}
\vspace{-1.5em}
\centering
\small
\setlength{\tabcolsep}{2pt}
\caption{Revision utility is polarized. Buckets count how many of the five runs of revisions improve over the initial unimodal forecast for each sample within our ID test set.}
\label{tab:cot_sft_revision_success_buckets}
\resizebox{\linewidth}{!}{
\begin{tabular}{@{}lrrrrrr@{}}
\toprule
 Validity  Bucket & 0/5 & 1/5 & 2/5 & 3/5 & 4/5 & 5/5 \\
\midrule
Percent & 29.17\% & 9.92\% & 7.76\% & 7.96\% & 10.09\% & 35.10\% \\
\bottomrule
\end{tabular}
}
\vspace{-1.0em}
\end{wraptable}

Table~\ref{tab:cot_sft_revision_success_buckets} shows why clean-sample filtering is risky. Many examples are easy, but the ambiguous 1/5 and 2/5 buckets already account for over 17\% of the data, and 0/5 non-revisable cases account for 29.17\%. Dropping them would remove precisely the cases useful to learn robust context use and  greatly reduce data utilization.

\subsubsection{Our CoT-SFT Design with No Human in the Loop}
\label{subsec:sft_data}
We would like to start from a \timex example and have a trace that has a good reasoning strategy for a revision and then the revised forecast that follows from that strategy. The gold standard for this would be human feedback but that is prohibitively expensive. Therefore we employ a high-quality and cost-effective frontier LLM - Gemini-3.1-Flash-Lite in particular - as a reasoning trace generator.

We start by appending all \timex training examples with their base TimesFM-2.5 forecasts. The examples are then prepared as prompts to query the trace generator for 5 independent traces per example. These traces are mandated to contain both the reasoning and a final candidate forecast.

Our proposed next step is to apply a filter to the initial SFT corpus to (1) encourage diversity and (2) preserve hard cases. For trace $j$ on example $i$, let $\hat y_{ij}$ be the candidate forecast and $\trace_{ij}$ its native decision trace. We classify a candidate as \textit{effective} by comparing it against the forecast prior:
\begin{equation}
  u_{ij}
  =
  \mathbb{1}\!\left[
  \mathrm{MAE}(\hat y_{ij}, y_i)
  <
  \mathrm{MAE}(\tilde{y}_i, y_i)
  \right].
  \label{eq:verified_effective}
\end{equation}
Among effective candidates, we retain at most $K=3$ traces with the lowest candidate MAE:
\begin{equation}
  \mathcal{S}_i =
  \operatorname*{TopK}_{j:u_{ij}=1}
  \left(-\mathrm{MAE}(\hat y_{ij},y_i)\right),
  \qquad K=3.
  \label{eq:topk_verified}
\end{equation}
If no trace improves over the forecast prior, we insert one fallback target $(\trace_i^{\mathrm{fb}}, \tilde{y}_i)$, where the trace states that the context does not support a reliable revision and the forecast remains the forecast prior. When none of the revisions are an improvement over the forecast prior it is a learning opportunity for the policy $\pi_{\theta}$ to leave the base forecast unchanged under such situations.

The resulting filtered SFT dataset is
\begin{equation}
\mathcal{D}_{\mathrm{SFT}} =
\bigcup_i
\begin{cases}
\{(x_i,c_i,\tilde{y}_i,\trace_{ij},\hat y_{ij}) : j\in\mathcal{S}_i\}, & |\mathcal{S}_i|>0,\\
\{(x_i,c_i,\tilde{y}_i,\trace_i^{\mathrm{fb}},\tilde{y}_i)\}, & |\mathcal{S}_i|=0.
\end{cases}
\label{eq:sft_dataset}
\end{equation}

\subsubsection{Empirical Evidence}
We show the empirical validation of our design choices when  applying \method to training an LLM reviser, whose details we later present in Section~\ref{sec:post_training_recipe}.
\begin{table}[h]
\centering
\caption{CoT-SFT data selection designs analysis.}
\label{tab:cot_sft_data_selection_recipes}
\begin{tabular}{@{}lrrrrrr@{}}
\toprule
CoT-SFT data selection & nMSE $\downarrow$ & p50 $\downarrow$ & p75 $\downarrow$ & p90 $\downarrow$ & p95 $\downarrow$ & p99 $\downarrow$ \\
\midrule
High-validity (4/5-5/5) buckets & 0.772 & 0.576 & 0.946 & 1.292 & 1.764 & 3.704 \\
All revisable buckets  (1/5-5/5)  & 0.665 & 0.558 & \textbf{0.862} & 1.229 & 1.584 & 1.808 \\
All Buckets (0/5-5/5;fallback traces for 0/5)& \textbf{0.650} & \textbf{0.525} & \textbf{0.862} & \textbf{1.187} & \textbf{1.267} & \textbf{1.742} \\
\midrule
\timesfm-2.5 & 0.729 & 0.600 & 0.956 & 1.303 & 1.471 & 2.616 \\
\bottomrule
\end{tabular}
\vspace{-1.0em}
\end{table}
\begin{table}[h]
\centering
\newcommand{\cotSftGtVerifierTableFont}{\small}
\newcommand{\cotSftTraceSelectionTableFont}{\small}
\begin{minipage}[t]{0.43\linewidth}
\vspace{0pt}
\centering
\cotSftGtVerifierTableFont
\setlength{\tabcolsep}{2pt}
\renewcommand{\arraystretch}{0.95}
\caption{Ground truth (GT) is a verifier, not an imitation target.}
\label{tab:cot_sft_gt_verifier}
\label{tab:sft_hindsight_vs_forecast_trace}
\begin{tabular*}{\linewidth}{@{\extracolsep{\fill}}lrr@{}}
\toprule
CoT-SFT target & nMAE & nMSE \\
\midrule
Direct GT fitting & 0.881 & 1.266 \\
GT hinted traces & 0.789 & 0.748 \\
Forecast-time traces & 0.781 & 0.716 \\
\midrule
\timesfm-2.5 & 0.788 & 0.729 \\
\bottomrule
\end{tabular*}
\end{minipage}
\hfill
\begin{minipage}[t]{0.53\linewidth}
\vspace{0pt}
\centering
\cotSftTraceSelectionTableFont
\setlength{\tabcolsep}{2pt}
\renewcommand{\arraystretch}{0.95}
\caption{Trace selection trades off SFT quality and token cost.}
\label{tab:cot_sft_trace_selection}
\begin{tabular*}{\linewidth}{@{\extracolsep{\fill}}lrrr@{}}
\toprule
Trace selection & nMAE $\downarrow$ & nMSE $\downarrow$ & Tokens $\downarrow$ \\
\midrule
Top1 effective & 0.781 & 0.716 & 12.4M \\
Random3 of 5 & 0.767 & 0.684 & 53.6M \\
Top3 effective & 0.761 & 0.665 & 32.6M \\
All effective & 0.758 & 0.665 & 47.2M \\
\midrule
\timesfm-2.5 & 0.788 & 0.729 & -- \\
\bottomrule
\end{tabular*}
\end{minipage}
\vspace{-0.8em}
\end{table}

\underline{\textit{The downside of fitting ground truth.}}
Table~\ref{tab:cot_sft_gt_verifier} supports using ground truth as a verifier rather than an imitation target. Direct ground-truth fitting is much worse than the \timesfm-2.5 prior, and ground-truth-guided hindsight still trails the prior in nMSE. In contrast, forecast-time traces improve over the prior on both nMAE and nMSE.

\underline{\textit{The value of hard cases.}}
Table~\ref{tab:cot_sft_data_selection_recipes} shows that high-validity filtering improves some typical quantiles but harms the tail: the high-positive recipe reaches a p99 nMSE of 3.704, worse than the \timesfm prior. Adding all revisable cases reduces the mean nMSE to 0.665 and the p99 nMSE to 1.808. Adding fallback supervision for the hardest cases, namely the 0/5 bucket, further improves the mean nMSE to 0.650 and the p99 nMSE to 1.742, outperforming the prior in both mean and tail behavior.

\underline{\textit{The superiority of diverse reasoning traces.}}
Table~\ref{tab:cot_sft_trace_selection} reports input plus output tokens under the Gemma-3-4B tokenizer and shows that a single selected path is not enough. Raw revising is worse than the prior, and Top1 effective gives only a modest gain. Multiple traces improve substantially: Top3 effective almost matches All effective while reducing SFT tokens from 47.2M to 32.6M, a 31.0\% reduction.

%% file: Sections/05a_main_results.tex
\section{Our Post-Training Recipe for Multimodal Time-Series Forecasting}
\label{sec:post_training_recipe}

Based on the findings in the rethinking sections, we propose a two-stage recipe for training an LLM to revise unimodal forecasts. In the CoT-SFT stage, we train the model to imitate diverse reasoning traces that reflect multiple plausible future trajectories. We suggest to include all validity buckets with fallback for unrevisable cases, so that the model learns when and how to use textual context robustly. In the RL stage, we use improvement over the unimodal prior, rather than absolute accuracy, to reduce reward collapse and encourage effective revision behavior.

Unless otherwise specified, we use Gemini-3.1-Flash-Lite as the reasoning-trace generator and Gemma-3-4B as the target revisor LLM.

\section{Main Results}
We benchmark our proposed post-training recipe against the following modern multimodal forecasting solutions.
\begin{itemize}
    \item \textbf{TSFMs:} We query unimodal time series foundation models that do not use textual context, including TimesFM-2.5~\citep{das2023decoder}, Moirai-2~\citep{woo2024unified}, and Sundial~\citep{liu2025sundial}.
    \item \textbf{Supervised methods:} We use advanced supervised models, TTS~\citep{li2026language}, which fuse LMs with TSF models, with 5 specific model combinations.
    \item \textbf{Zero-shot LLMs:} We directly prompt frontier LLMs with the numerical history and textual context to perform multimodal forecasting.
    \item \textbf{Training-free LLM--TSFM integrations:} We consider two common ways to combine pretrained LLMs and TSFMs without post-training: ensembling (ENS), which averages forecasts from an LLM and a TSFM, and revision (REV), which uses zero-shot prompting to let an LLM revise a TSFM forecast.
    \item \textbf{TSLMs:} Finally, we query continuously pretrained time series language models, including TimeOmni-1-7B~\citep{guan2025timeomni}, Aurora~\citep{wu2025aurora}, and ChatTS-8B\citep{xie2025chatts}.
\end{itemize}

The knowledge-cutoff dates of all considered LLMs are before our dataset cutoff, Jan.~30, 2025, which prevents information leakage during evaluation. Our experiments were conducted on a dual-A100 (80GB) server. We reran the experiments three times to report the average.

\begin{table}[ht]

\centering
\caption{Main results on \timex, combining ID (88 variables) and OOD (11 variables) test splits. Metrics are computed per example then averaged. Lower is better. Bold font marks the full \method recipe and the best metric per column.}
\label{tab:main_results_iid_ood_combined}
{\small
\setlength{\tabcolsep}{2pt}
\renewcommand{\arraystretch}{0.70}
\resizebox{0.98\textwidth}{!}{
\begin{tabular}{@{}llrrrr@{}}
\toprule
Group & Method & ID nMAE $\downarrow$ & ID nMSE $\downarrow$ & OOD nMAE $\downarrow$ & OOD nMSE $\downarrow$ \\
\midrule
\multirow{9}{*}{\textit{Supervised}} & TTS (iTransformer + GPT2) & 1.289 & 2.980 & 1.188 & 1.508 \\
 & TTS (FiLM + GPT2) & 1.441 & 3.580 & 1.637 & 2.186 \\
 & TTS (DLinear + GPT2) & 1.513 & 3.464 & 1.836 & 2.766 \\
 & TTS (PatchTST + GPT2) & 1.531 & 6.212 & 1.705 & 3.819 \\
 & TTS (Crossformer + GPT2) & 2.037 & 6.300 & 3.347 & 19.938 \\
\midrule
\multirow{18}{*}{\textit{Training-free}} & ENS (GPT-5 + \timesfm) & 0.760 & 0.791 & 0.769 & 0.660 \\
 & ENS (Qwen-3-VL-32B + \timesfm) & 0.795 & 0.753 & 0.901 & 1.083 \\
 & ENS (GPT-OSS-120B + \timesfm) & 0.794 & 1.154 & 0.857 & 0.787 \\
 & ENS (MiMo-V2-Flash + \timesfm) & 0.818 & 1.237 & 0.863 & 0.739 \\
 & ENS (Gemma-4-31B + \timesfm) & 0.802 & 1.925 & 0.828 & 0.801 \\
 & ENS (Gemini-3.1-Pro + \timesfm) & 0.807 & 2.715 & \textbf{0.742} & 0.637 \\
 & ENS (Gemini-2.5-Flash + \timesfm) & 0.944 & 3.905 & 0.856 & 0.852 \\
 & REV (GPT-5 + \timesfm) & 0.783 & 0.725 & 0.836 & 0.737 \\
 & REV (GPT-OSS-120B + \timesfm) & 0.794 & 0.787 & 0.950 & 1.331 \\
 & REV (Qwen-3-VL-32B + \timesfm) & 0.915 & 1.026 & 1.356 & 4.222 \\
 & REV (Gemini-2.5-Flash + \timesfm) & 1.151 & 6.716 & 0.967 & 1.380 \\
 & REV (Gemini-3.1-Pro + \timesfm) & 0.949 & 9.375 & 0.787 & 0.659 \\
 & REV (Gemma-3-4B + \timesfm) & 0.877 & 0.878 & 0.894 & 0.883 \\
\midrule
\multirow{3}{*}{\textit{TSLMs}} & TimeOmni-1-7B & 1.088 & 1.417 & 0.777 & 0.673 \\
 & Aurora & 1.084 & 1.387 & 1.025 & 1.113 \\
 & ChatTS-8B & 1.189 & 1.689 & 1.117 & 1.394 \\
\midrule
\multirow{11}{*}{\textit{LLMs}} & GPT-5 & 0.843 & 1.299 & 0.846 & 0.809 \\
 & Qwen-3-VL-32B & 0.947 & 1.122 & 1.163 & 2.395 \\
 & GPT-OSS-120B & 0.930 & 2.762 & 1.027 & 1.177 \\
 & MiMo-V2-Flash & 0.969 & 3.039 & 1.037 & 1.068 \\
 & Gemma-4-31B & 0.955 & 5.717 & 0.999 & 1.239 \\
 & Gemini-3.1-Pro & 0.917 & 8.890 & 0.771 & 0.702 \\
 & Gemini-2.5-Flash & 1.261 & 13.445 & 1.056 & 1.447 \\
 & Gemma-3-4B & 1.486 & 11.434 & 1.306 & 1.944 \\
\midrule
\multirow{3}{*}{\textit{TSFMs}} & \timesfm-2.5 & 0.788 & 0.729 & 0.777 & 0.673 \\
 & Moirai-2.0 & 0.800 & 0.731 & 0.942 & 0.905 \\
 & Sundial & 1.121 & 3.144 & 1.057 & 1.211 \\
\midrule
\multirow{2}{*}{\textsc{PostTime}} & SFT (Gemma-3-4B + \timesfm) & 0.739 & 0.650 & 0.772 & 0.623 \\
 & \textbf{SFT+RL (Gemma-3-4B + \timesfm)} & \textbf{0.738} & \textbf{0.638} & 0.746 & \textbf{0.597} \\
\bottomrule
\end{tabular}
}
}
\end{table}

Table~\ref{tab:main_results_iid_ood_combined} demonstrates the main benchmark results, while extended results involving more baselines can be found in Appendix~\ref{sec:extended_results}. Despite its minimal 4B size, the \method Gemma-3-4B model outperforms on both the ID and the OOD fronts almost all other baseline methods, some of which involve the most advanced commercial LLMs and cutting-edge TSLMs.

In terms of measuring the revising power, the \method Gemma-3-4B improves the nMAE and the nMSE of base TimesFM-2.5 forecasts by 6.38\% and 12.43\% on the ID test split, and 3.93\% and 11.24\% on the OOD test split respectively. Besides, both the SFT and SFT+RLVR Gemma-3-4B achieve 100\% success rate in terms of always generating a valid final forecast during our benchmarking.

These benchmark results empirically support our choice of building a LLM reviser, and the decisions made regarding SFT data collection and RL reward design.

%% file: Sections/06_further_analysis.tex
\section{Further Analysis}
\label{sec:further_analysis}

The main results show that \method improves over strong TSFMs on both ID and OOD evaluations. We next ask whether these gains depend on a particular trace generator or a particular base LLM.

\subsection{Generalizing \method to other \tracegen}
\label{subsec:rq_annotation_llm}

In our main results we had used Gemini-3.1-Flash-Lite as the \tracegen. Here we show that our findings do not rely on that choice. Table~\ref{tab:annotation_llm_robustness} compares two commercial LLMs as \tracegen s: Gemini-3.1-Flash-Lite (G3.1) and Gemini-3-Flash (G3) .Direct revising (without post-training) is highly unstable, especially on ID nMSE: G3.1 reaches 2.966 and G3 reaches 7.349. After effective-trace selection and SFT, however, both \tracegen s produce substantially stronger models. With G3.1 traces, \method SFT reaches 0.739 nMAE / 0.650 nMSE on ID and 0.772 nMAE / 0.623 nMSE on OOD. With G3 traces, it still reaches 0.760 / 0.695 on ID and improves OOD nMAE to 0.746.

This suggests that in both cases our SFT recipe helps the base LLM move beyond the zero-shot revising capability of the \tracegen, thus highlighting the robustness of our recipe to this choice. At the same time, the \tracegen~  is not irrelevant: G3.1 traces are stronger on ID, whereas G3 traces are better on OOD nMAE. 

\begin{table}[ht]
  \centering
  \caption{Impact of two \tracegen s: Gemini-3.1-Flash-Lite (G3.1) vs. Gemini-3-Flash (G3).}
  \label{tab:annotation_llm_robustness}
  \begin{tabular}{lcccc}
    \toprule
    Method & ID nMAE $\downarrow$ & ID nMSE $\downarrow$ & OOD nMAE $\downarrow$ & OOD nMSE $\downarrow$ \\
    \midrule
    Zero-shot REV (G3.1)       & 0.820 & 2.966 & 0.827 & 0.752 \\
    \method SFT on G3.1 Traces & 0.739 & 0.650 & 0.772 & 0.623 \\
    Zero-shot REV (G3)         & 0.950 & 7.349 & 0.801 & 0.698 \\
    \method SFT on G3 Traces   & 0.760 & 0.695 & 0.746 & 0.671 \\
    \bottomrule
  \end{tabular}
\end{table}

\subsection{Generalizing \method to other base LLMs}
\label{subsec:rq_backbone_llm}

\begin{table}[ht]
  \centering
  \caption{Impact of changing the base LLM. \method delivers the same performance progression when applied to Qwen3-4B.}
  \label{tab:qwen_backbone_robustness}
  \begin{tabular}{lcccc}
    \toprule
    Method & ID nMAE $\downarrow$ & ID nMSE  $\downarrow$ & OOD nMAE $\downarrow$ & OOD nMSE $\downarrow$ \\
    \midrule
    Qwen3-4B       & 1.054 & 1.400 & 1.625 & 3.029 \\
    \method SFT    & 0.760 & 0.739 & 0.766 & 0.634 \\
    \method SFT+RL & 0.747 & 0.651 & 0.745 & 0.609 \\
    \bottomrule
  \end{tabular}
\end{table}

Table~\ref{tab:qwen_backbone_robustness} replaces the base LLM Gemma-3-4B with Qwen3-4B. The initial Qwen3-4B model is a weak reviser, with 1.054 nMAE / 1.400 nMSE on ID and 1.625 nMAE / 3.029 nMSE on OOD. CoT-SFT provides the main capability jump, reducing the OOD error to 0.766 nMAE / 0.634 nMSE. RL then further improves the already trained model to 0.747 / 0.651 on ID and 0.745 / 0.609 on OOD.

While we observe the monotonic progression from initial to SFT to SFT+RL on Gemma-3-4B, observing this same progression pattern on Qwen3-4B supports the view of \method as a recipe-level post-training approach rather than a base LLM specific result.

\subsection{Falling back to \tsfm priors}
\label{subsec:rq_fallback_behavior}

A post-trained reviser should have the ability to judge what parts of the context are relevant for the forecasting task. The textual context can be weak, stale, or only loosely related to the prediction window, therefore a robust reviser should also know when to preserve the base \tsfm forecast. We inspect inference-time fallbacks, where the generated forecast equals the \tsfm prior (detailed definitions are provided in Appendix~\ref{app:inference_fallback_metrics}).

\begin{wraptable}{r}{0.43\textwidth}
  \vspace{-1.2em}
  \centering
  \caption{Fallback rate after post-training.}
  \label{tab:inference_fallback_behavior}
  \footnotesize
  \setlength{\tabcolsep}{4pt}
  \begin{tabular}{@{}lrr@{}}
    \toprule
    Split & SFT & SFT+RL \\
    \midrule
    ID & 22.90\% & 21.76\% \\
    OOD & 23.11\% & 17.45\% \\
    \bottomrule
  \end{tabular}
  \vspace{-1.0em}
\end{wraptable}

Table~\ref{tab:inference_fallback_behavior} shows that fallback remains below roughly one quarter of all cases, so both model after either SFT or SFT+RL stages usually attempt a revision rather than preserving the prior. RL makes the model revise more often, especially under distribution shift: on OOD cases, fallback drops from 23.11\% to 17.45\%. However, the fallback rate is not vanishingly small i.e the models do learn to not intervene in some case. Now lets analyze these cases in more detail.

\begin{table}[t]
\centering
\small
\caption{Characteristics of final SFT+RL decisions. Fallback cases have weaker textual evidence and smoother histories; lower rMAFD means less nonstationary historical trajectories.}
\label{tab:inference_fallback_characteristics}
\setlength{\tabcolsep}{5pt}
\begin{tabular}{@{}lrrr@{}}
\toprule
Action & Context words & Closest event gap & Nonstationarity (rMAFD) \\
\midrule
Fallback & 1.2k & 21.0 days & 0.007 \\
Revision & 1.3k & 15.0 days & 0.013 \\
\bottomrule
\end{tabular}
\end{table}

Table~\ref{tab:inference_fallback_characteristics} asks where the final SFT+RL model still chooses not to intervene, pooling ID and OOD outputs. Fallback cases have shorter contexts, less temporally proximate event evidence, and lower rMAFD, indicating smoother histories (see Appendix~\ref{app:inference_fallback_metrics}). This provides further evidence that the post-training recipe indeed teaches the LLM to not intervene under some reasonable circumstances, for instance when the textual context is of low quality.

\section{Conclusions}

In this paper we introduced \method, a post-training recipe for building an LLM reviser to provide context-guided revision of a TSFM prior in order to solve multimodal time series forecasting tasks. We have provided detailed discussions regarding the design choices made when creating \method, and have empirically shown that applying \method to a Gemma-3-4B model revising TimesFM-2.5 priors achieves outstanding multimodal forecasting performance.

While \method focuses on the thesis that LLMs play the role of a textual reviser, there are many other potential interaction mechanisms between an LLM and a TSFM to together deliver a multimodal solution. Despite its limitations (see Appendix~\ref{app:limitation}), we believe \method is a good reference when considering the design choices for any of such interactions when post-training an LLM is required.

\section*{Acknowledgment}

This work was partly supported by the NSF (Expeditions CCF-1918770, CAREER IIS-2028586, Medium IIS-1955883, Medium IIS-2403240, Medium IIS-2106961), NIH (1R01HL184139), Meta, Modal.com and Dolby faculty gifts.

\newpage

%% file: Sections/A_recipe_implementation_details.tex
\newpage
\section{Limitations and Future Work}
\label{app:limitation}
We have demonstrated that \method represents a performant post-training recipe to make a base LLM an adept multimodal forecaster by potentially revising the forecasts of a TSFM after incorporating the textual context into its reasoning. We have also shown that the recipe is robust to the choice of various base LLMs and \tracegen s. However several options remain open for future work in this direction:

(1) Both the benchmark and the methods discussed here do not explore how the conclusions changes in the presence of even more modalities like images for instance the ones in financial reports that could be relevant for asset price forecasting. That is an interesting direction which is beyond the scope of this paper, even though the key steps in the recipe could be adapted to such a setting.

(2) There could be many other ways in which the LLM could be used to revise a forecast. For instance, in this work we do not explore modifying the forecasts through code execution or other forms of tool use. This is partly because the revising setting forms a clean test bed for our ablation studies and the salient points in the design of our recipe can also be applied under more general environments in future efforts.

(3) Future work could also look into building systems which can query the web or knowledge graph to gather more contexts when the LLM decides that the existing is not good enough for the revising task.

(4) Future work also includes extending to other time-series reasoning tasks~\citep{liu-etal-2025-picture}, as well as general event forecasting tasks~\citep{liu2601futurex}. But these tasks are beyond the scope of time-series forecasting.
\section{Related Work}
\label{sec:related_work}

\subsection{Time-Series Foundation Models}
\label{subsec:rw_tsfm}

Time-series foundation models (TSFMs) have recently become strong zero-shot forecasters for numerical time series. Models such as TimesFM~\citep{das2023decoder}, Moirai~\citep{woo2024unified}, Chronos-2~\citep{ansari2025chronos}, and Sundial~\citep{liu2025sundial} are trained on large collections of temporal data and can produce competitive forecasts from numerical histories alone. These models are especially useful when no task-specific supervised training data~\citep{kamarthi2024large} are available, which makes them a natural starting point for broad forecasting benchmarks.
\subsection{Multimodal Time-Series Forecasting and Benchmarks}
\label{subsec:rw_multimodal_tsf}

Real-world forecasting often depends on information beyond the observed numerical history~\citep{liu2025can}. Early multimodal forecasting benchmarks start to expose this need by pairing time series with textual context. For example, CiK~\citep{ williams2025context} contains real-world time series with manually crafted textual contexts that are critical for forecasting. Time-MMD~\citep{liu2024time} further provides a context-enriched benchmark constructed through keyword-based web search, covering nine domains with one variable per domain. More recently, \timex~\citep{liu2026rethinking} expands this direction by incorporating multiple types of forecast-time context, including metadata, calendar information, covariates, and time-stamped events. It also identifies leakage issues in existing multimodal forecasting evaluations and expands the benchmark scale. Our work builds on the public \timex dataset agent and extends its original evaluation suite into a post-training corpus for learning context-conditioned forecast revision.

Several recent methods train or adapt models to consume both time-series and language inputs. TTS~\citep{li2026language} studies language-time-series paired data and supervised multimodal forecasting. Aurora~\citep{wu2025aurora} studies generative multimodal time-series forecasting through multimodal pretraining. ChatTS~\citep{xie2025chatts} aligns time series with LLMs through synthetic data for temporal understanding and reasoning, while Time-MQA~\citep{kong2025timemqa} studies multi-task question answering with time-series context. More recent reasoning-oriented systems, such as Time-R1~\citep{liu2025timer1} and TimeOmni~\citep{guan2025timeomni}, further introduce CoT-style supervision and RL to improve time-series reasoning ability.

\section{Implementation Details for the Post-Training Recipe}
\label{app:recipe_implementation_details}

\subsection{Forecasting Suite Construction}
\label{app:forecasting_suite_construction}

{\color{black}
The post-training data are derived from a \timex forecasting suite rather than from standalone text-response annotations. The \timex dataset agent aligns numerical time-series values with forecast-time multimodal context over 2022--2025. Each forecasting instance stores the historical values, requested prediction timestamps, context available at the forecast origin, the initial forecast from the time-series foundation model, and the realized future. The realized future is used only for candidate verification, SFT target selection, reward computation, and evaluation; it is never included in the student prompt at forecast time.
}

\begin{table}[h]
\centering
\small
\caption{\textcolor{black}{Data split protocol for the forecasting suite. Splits are defined by prediction windows, so held-out evaluation windows are not exposed during post-training.}}
\label{tab:app_forecasting_suite_splits}
\resizebox{0.98\textwidth}{!}{
\color{black}
\begin{tabular}{@{}llll@{}}
\toprule
Split & Variables & Prediction-window rule & Use \\
\midrule
Post-training & Seen variables & Starts after Jan.~1, 2023; ends before Feb.~1, 2025 & Proposal, SFT, RL \\
IID evaluation & 88 seen variables & Starts after Jan.~30, 2025 & Held-out IID test \\
OOD evaluation & 11 unseen variables & Starts after Jan.~30, 2025 & Out-of-domain test \\
\bottomrule
\end{tabular}
}
\end{table}

{\color{black}
This protocol separates three possible leakage channels. The prompt context is restricted to information available at the forecast origin, the post-training corpus does not contain held-out evaluation target windows, and OOD evaluation additionally changes the variable set while retaining the later forecast horizon.

\paragraph{ID variables.}
The ID evaluation uses the following seen-variable set:
\begin{sloppypar}
\small\ttfamily
\detokenize{food_wine_festivals, national_football_league, polypropylene_cny_t, copper_usd_lbs, beef_brl_kg, salmon_nok_kg, palladium_usd_t_oz, heavy_rainfall, naphtha_usd_t, global_warming, gold_usd_t_oz, canola_cad_t, animal_rescue, major_league_baseball, usdtomxn_exchangerate, indium_cny_kg, heatwave, gallium_cny_kg, usdtocad_exchangerate, nickel_usd_t, unemployment_rate, butter_eur_t, drug_overdose, urea_usd_t, cobalt_usd_t, pest_control, polyvinyl_cny_t, film_festivals, lumber_usd_1000_board_feet, usdtokrw_exchangerate, molybdenum_cny_kg, infectious_disease, lithium_cny_t, beekeeping, ethanol_usd_gal, corn_usd_bu, uranium_usd_lbs, water_scarcity, usdtoaud_exchangerate, silver_usd_t_oz, sugar_usd_lbs, rubber_usd_cents_kg, cost_of_living, cocoa_usd_t, propane_usd_gal, magnesium_cny_t, platinum_usd_t_oz, coffee_usd_lbs, music_festivals, milk_usd_cwt, oat_usd_bu, potatoes_eur_100kg, drought, national_basketball_association, pet_adoption, wheat_usd_bu, carbon_emissions, cheese_usd_lbs, inflation, neodymium_cny_t, brent_usd_bbl, usdtobrl_exchangerate, rhodium_usd_t_oz, barley_inr_t, food_safety, esports, diabetes, usdtogbp_exchangerate, poultry_brl_kgs, meta_platforms, deforestation, aluminum_usd_t, soybeans_usd_bu, invasive_species, hiv_aids, diammonium_usd_t, usdtohkd_exchangerate, animal_migration, coal_usd_t, marine_pollution, gasoline_usd_gal, rapeseed_eur_t, wool_aud_100kg, lead_usd_t, tin_usd_t, cotton_usd_lbs, methanol_cny_t, climate_change}
\end{sloppypar}

\paragraph{OOD variables.}
The OOD evaluation uses the following held-out variable set:
\begin{sloppypar}
\small\ttfamily
\detokenize{comic_con, zinc_usd_t, biodiversity, steel_cny_t, usdtoinr_exchangerate, endangered_species, rice_usd_cwt, germanium_cny_kg, air_pollution, tellurium_cny_kg, polyethylene_cny_t}
\end{sloppypar}

\paragraph{Version-C construction note.}
The main benchmark and main results use the 88-variable seen set above. Some Version-C trusted SFT construction artifacts contain 87 variables because the variable \texttt{food\_wine\_festivals} did not have trusted annotation in that construction branch. This construction detail does not change the main IID evaluation scope.
}

\subsection{Example \timex Forecasting Instance}
\label{app:timex_data_demo}

\begin{figure}[h]
  \centering
  \includegraphics[width=0.86\linewidth]{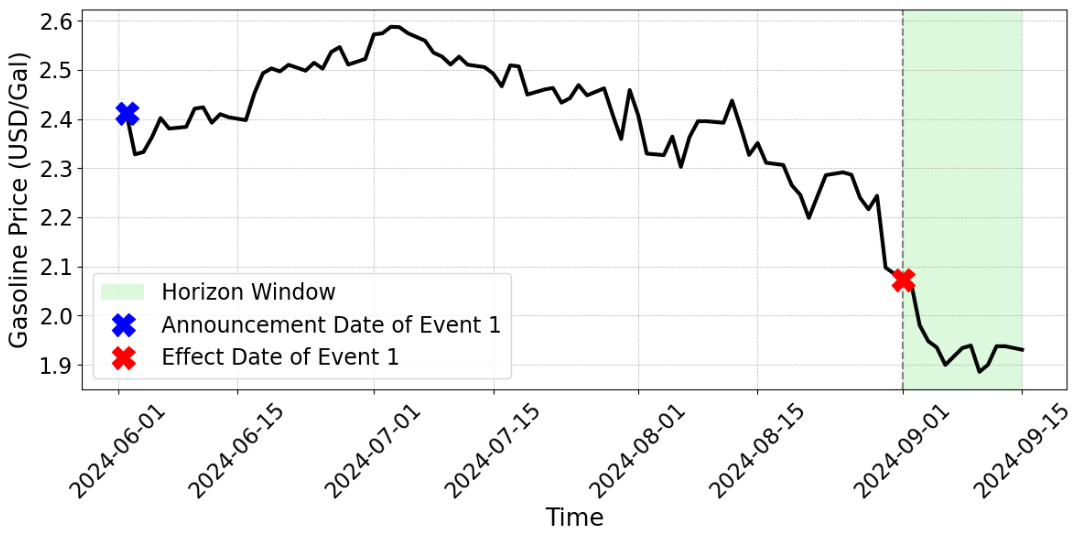}
  \caption{\textcolor{black}{Numerical example from the \textsc{GAS\_PRICE} variable in the \timex data demo.}}
  \label{fig:app_eval_gas_price}
\end{figure}

{\color{black}
We provide below an example from \timex corresponding to the \textsc{GAS\_PRICE} time series, following the dataset demo prepared in the original \timex paper. The example illustrates how numerical history, metadata, calendar information, covariates, and event context are aligned to the same forecast origin.

\textbf{Time series}: See Fig.~\ref{fig:app_eval_gas_price}. \textbf{Metadata}: This time series records gasoline price (USD/GAL) in the Commodity Price domain, with daily frequency. Prediction target period: from 2024-09-01 to 2024-09-15. \textbf{Date}: Upcoming holidays in the prediction window: Labor Day (2024-09-02). \textbf{Covariates}: from 2024-06-01 to 2024-08-31: Brent Crude Oil (USD/BBL) reached a maximum value of 87.43 on July 4 and then showed an overall downward trend, alongside other related energy and commodity covariates. \textbf{Events}: On June 2, 2024, OPEC+ agreed to extend deep oil output cuts; the cut of 2.2 million bpd would be extended until September 2024, after which it would be gradually phased out.
}

\subsection{Prompt and Output Format}
\label{app:prompt_output_format}

{\color{black}
We use two prompt families. The direct-forecast prompt asks the LLM to forecast from the history and context alone. The revising prompt additionally provides an initial forecast from a time-series foundation model and asks the LLM to revise it only when the context supports a reliable change.
}

\begin{figure*}[h]
  \centering
  \begin{adjustbox}{max totalsize={\textwidth}{0.82\textheight}}
    \begin{tcolorbox}
      \begin{lstlisting}[style=redprompt]
You are an expert forecaster.

Here is some context about the task. Please consider this information when making the forecast:
<context>
{background information, metadata, calendar signals, covariates, recent events, and other forecast-time context}
</context>

Here is the historical time series in (timestamp, value) format:
<history>
{historical data points}
</history>

Requested timestamps:
{future timestamps to forecast}

Please produce the forecast for the requested timestamps.
Return the forecast in (timestamp, value) format between <forecast> and </forecast> tags.
Do not include any other information outside the tags.

Example format:
<forecast>
(2024-01-01 00:00:00, 123.45)
(2024-01-02 00:00:00, 125.67)
</forecast>
      \end{lstlisting}
    \end{tcolorbox}
  \end{adjustbox}
  \caption{\textcolor{black}{Direct forecast prompt template. The prompt provides context and history, but no initial TSFM forecast.}}
  \label{fig:app_direct_forecast_prompt}
\end{figure*}

\begin{figure*}[h]
  \centering
  \begin{adjustbox}{max totalsize={\textwidth}{0.86\textheight}}
    \begin{tcolorbox}
      \begin{lstlisting}[style=redprompt]
You are an expert forecaster tasked with refining a statistical forecast by applying contextual reasoning.

Follow this workflow:
1. Study the contextual information and historical values to identify relevant drivers, constraints, or anomalies.
2. Compare the initial forecast against these signals and decide whether the context supports a reliable change.
3. Summarize your reasoning inside <analysis></analysis>, explaining any adjustments or why the values should stay the same.
4. Output the final forecast inside <forecast></forecast> using (timestamp, value) lines in the same order as the requested timestamps.

<context>
{background information, metadata, calendar signals, covariates, recent events, and other forecast-time context}
</context>

<history>
{historical data points}
</history>

<initial_forecast>
{initial forecast from the time-series foundation model}
</initial_forecast>

Requested timestamps:
{future timestamps to forecast}

Example format:
<analysis>
- Context driver A points to stronger demand in the first half of the horizon.
- I adjust the early timestamps upward while keeping later timestamps close to the initial forecast.
- If the context is weak or conflicting, I keep the initial forecast unchanged.
</analysis>
<forecast>
(2024-01-01 00:00:00, 123.45)
(2024-01-02 00:00:00, 125.67)
</forecast>
      \end{lstlisting}
    \end{tcolorbox}
  \end{adjustbox}
  \caption{\textcolor{black}{Revising prompt template. The prompt provides the initial TSFM forecast, and preserving that forecast is a valid action when context is not sufficiently reliable.}}
  \label{fig:app_revising_prompt}
\end{figure*}

{\color{black}
For revising, the forecast block must follow the requested timestamp order. Fallback examples use the same format, but the forecast equals the initial forecast and the analysis states that the available context does not justify modifying the numerical prior.
}

\subsection{Top-3/Fallback Dataset Construction}
\label{app:top3_fallback_construction}

Candidate proposals are verified independently for each source example. A proposal is retained if its MAE is lower than the MAE of the base forecast for the same target window, as in Eq.~\ref{eq:verified_effective}. If at least one proposal is effective, we retain up to three candidates with the lowest MAE. If no proposal is effective, we add exactly one fallback example. The test split is never used for SFT construction.

\begin{table}[h]
\centering
\small
\caption{Scale of the top-3/fallback SFT corpus. Token counts are measured with the Gemma-3-4B-PT tokenizer over prompt plus assistant answer.}
\label{tab:app_sft_data_scale}
\begin{tabular}{@{}lr@{}}
\toprule
Item & Value \\
\midrule
Training source samples & 4,587 \\
SFT rows after top-3/fallback & 10,073 \\
Verified intervention rows & 8,798 \\
Fallback rows & 1,275 \\
Prompt tokens & 57.94M \\
Answer tokens & 4.58M \\
\bottomrule
\end{tabular}
\end{table}

Before filtering, the all-five proposal pool contains $38{,}160$ candidate traces over successful examples. This pool is used to expose the verification stage to diverse teacher interventions; the SFT corpus is the compact subset retained after utility filtering and fallback insertion.

\subsection{SFT Training Details}
\label{app:sft_training_details}

\begin{table}[h]
\centering
\small
\caption{SFT hyperparameters for the verified intervention student.}
\label{tab:app_sft_hyperparams}
\begin{tabular}{@{}ll@{}}
\toprule
Hyperparameter & Value \\
\midrule
Base model & Gemma-3-4B-PT \\
Training stage & SFT \\
Finetuning type & LoRA \\
LoRA rank / alpha / dropout & 8 / 16 / 0.05 \\
LoRA target & all modules \\
Template & Gemma-3 \\
Cutoff length & 8192 \\
Epochs & 3.0 \\
Per-device train batch size & 1 \\
Gradient accumulation steps & 16 \\
Learning rate & $1.3\times 10^{-4}$ \\
\bottomrule
\end{tabular}
\end{table}

\subsection{Sample-Normalized Trace-and-Forecast SFT Loss}
\label{app:sample_normalized_sft_loss}

Each retained SFT row contains a prompt $(x_m,c_m,y_m^0)$ and a target response $z_m=(r_m,\bar y_m)$. The trace $r_m$ is serialized inside the analysis span, and $\bar y_m$ is serialized inside the forecast span. Let $\mathcal{T}_m^r$ and $\mathcal{T}_m^y$ denote their supervised positions and let $\mathcal{T}_m=\mathcal{T}_m^r\cup\mathcal{T}_m^y$. The row loss is the mean next-token cross entropy over this full supervised response:
\begin{equation}
\label{eq:sample_normalized_sft_loss_main}
  \ell_m(\theta)
  =
  \frac{1}{|\mathcal{T}_m|+\epsilon}
  \sum_{t\in\mathcal{T}_m}
  \mathrm{CE}_{m,t}(\theta),
  \qquad
  \mathrm{CE}_{m,t}(\theta)
  =
  -\log p_\theta(z_{m,t}\mid x_m,c_m,y_m^0,z_{m,<t}).
\end{equation}
The dataset objective in Eq.~\ref{eq:sample_normalized_sft_loss_main} averages this row loss over retained SFT rows. The main implementation sets \texttt{analysis\_loss\_weight=1.0} and \texttt{forecast\_loss\_weight=1.0}, so no additional span reweighting is applied. The sample normalization prevents longer traces from receiving larger aggregate weight solely because they contain more tokens.

\subsection{RL Training Details}
\label{app:rl_training_details}

The RL stage uses the \impratio reward in Eq.~\ref{eq:impratio_reward}. We train with a group-based policy optimization setup and LoRA adapters.

\begin{table}[h]
\centering
\small
\caption{RL hyperparameters for baseline-relative intervention calibration.}
\label{tab:app_rl_hyperparams}
\begin{tabular}{@{}ll@{}}
\toprule
Hyperparameter & Value \\
\midrule
RL objective family & GRPO  \\
Reward & \impratio \\
Finetuning type & LoRA \\
LoRA rank / alpha / dropout & 8 / 32 / 0.05 \\
Max sequence length & 10240 \\
Max completion length & 1024 \\
Per-device train batch size & 4 \\
Gradient accumulation steps & 2 \\
Learning rate & $1.0\times10^{-5}$ \\
Weight decay & 0.01 \\
Epochs & 8 \\
Warmup ratio & 0.03 \\
Max gradient norm & 0.3 \\
Number of generations & 4 \\
Steps per generation & 5 \\
Temperature / top-$p$ / top-$k$ & 0.9 / 0.9 / 50 \\
KL beta & 0.04 \\
Clipping epsilon & 0.2 \\
\bottomrule
\end{tabular}
\end{table}

The reward is computed from parsed forecasts aligned to the requested timestamps. Invalid or unparsable completions receive zero reward, which preserves the strict output contract learned during SFT.

\subsection{Inference-Time Fallback Behavior Metrics}
\label{app:inference_fallback_metrics}

The fallback analysis in Section~\ref{subsec:rq_fallback_behavior} measures the behavior of trained SFT and SFT+RL models at inference time. This is distinct from the fallback examples inserted during SFT data construction in Appendix~\ref{app:top3_fallback_construction}.

\paragraph{Actual fallback.}
Actual fallback checks whether the generated \texttt{<forecast>} block equals the initial \tsfm forecast in the input prompt after the existing forecast parser and post-processing. This is the fallback metric reported in Table~\ref{tab:inference_fallback_behavior}: it measures whether the model's final forecasting action preserves the numerical prior.

\paragraph{Context words.}
Context length is the word count of the input \texttt{<context>} block. We use it as a simple proxy for the amount of textual evidence available to the model.

\paragraph{Closest event gap.}
We extract the forecast start date from the prompt and event dates from the recent-events section. The closest event gap is the minimum absolute distance in days between the forecast start date and any extracted event date. Larger values indicate that the nearest event evidence is less temporally proximate to the prediction window.

\paragraph{Historical nonstationarity.}
For a historical sequence $x_{1:T}$, we compute relative mean absolute first difference:
\begin{equation}
  \mathrm{rMAFD}(x_{1:T})
  =
  \frac{\frac{1}{T-1}\sum_{t=2}^{T}|x_t-x_{t-1}|}
       {\frac{1}{T}\sum_{t=1}^{T}|x_t|}.
\end{equation}
Lower rMAFD indicates a smoother, less nonstationary historical trajectory. In Table~\ref{tab:inference_fallback_characteristics}, fallback cases have lower rMAFD than revision cases, supporting the interpretation that no-op decisions are more common when the numerical prior is already stable.

\paragraph{Terminology.}
Inference-time fallback is a trained-model behavior: the model decides not to revise the initial \tsfm forecast. Training-data fallback is a construction rule used before SFT when no teacher candidate improves over the base forecast. Evaluator fallback is an engineering recovery used when generated outputs are invalid or unparsable. The RQ3 analysis uses only the first notion.

%% file: Sections/05_main_results.tex
\section{Extended Results}
\label{sec:extended_results}
See more methods and metric in Table~\ref{tab:main_results_iid} and ~\ref{tab:main_results_ood}.

\begin{table}[!p]
\centering
\caption{ID main results on 88 in-domain variables. The SFT-only \method variant is inserted immediately above the final RL-refined \method, which achieves the best average rank and improves over \timesfm-2.5 by 6.38\% nMAE and 12.43\% nMSE. Rows are grouped by the leftmost category column; lower is better, and bold marks the final \method and metric bests.}
\label{tab:main_results_iid}
{\scriptsize
\setlength{\tabcolsep}{2pt}
\renewcommand{\arraystretch}{0.70}
\resizebox{0.98\textwidth}{!}{
\begin{tabular}{@{}llrrr@{}}
\toprule
Group & Method &n MAE $\downarrow$ &n MSE $\downarrow$ & Avg. Rank $\downarrow$ \\
\midrule
\multirow{9}{*}{\textit{Supervised}} & TTS(iTransformer+GPT2) & 1.289 & 2.980 & 30.5 \\
 & TTS(FiLM+GPT2) & 1.441 & 3.580 & 34.5 \\
 & TTS(DLinear+GPT2) & 1.513 & 3.464 & 35.0 \\
 & TTS(Informer+GPT2) & 1.660 & 3.360 & 35.5 \\
 & TTS(Transformer+GPT2) & 1.758 & 5.341 & 38.5 \\
 & TTS(PatchTST+GPT2) & 1.531 & 6.212 & 38.5 \\
 & TTS(FEDformer+GPT2) & 2.224 & 10.530 & 44.0 \\
 & TTS(Autoformer+GPT2) & 2.373 & 12.238 & 45.5 \\
 & TTS(Crossformer+GPT2) & 2.037 & 6.300 & 40.5 \\
\midrule
\multirow{18}{*}{\textit{Training-free}} & ENS(GPT-5+\timesfm) & 0.760 & 0.791 & 6.0 \\
 & ENS(Qwen-3-VL-32B+\timesfm) & 0.795 & 0.753 & 7.5 \\
 & ENS(GPT-OSS-120B+\timesfm) & 0.794 & 1.154 & 10.0 \\
 & ENS(MiMo-V2-Flash+\timesfm) & 0.818 & 1.237 & 14.0 \\
 & ENS(Gemini-3.1-Flash-Lite+\timesfm) & 0.755 & 1.234 & 8.0 \\
 & ENS(Gemma-4-26B-A4B+\timesfm) & 0.827 & 1.634 & 17.0 \\
 & ENS(Gemma-4-31B+\timesfm) & 0.802 & 1.925 & 15.5 \\
 & ENS(Gemini-3.1-Pro+\timesfm) & 0.807 & 2.715 & 16.5 \\
 & ENS(Gemini-3-Flash+\timesfm) & 0.846 & 2.992 & 21.5 \\
 & ENS(Gemini-2.5-Flash+\timesfm) & 0.944 & 3.905 & 27.5 \\
 & REV(GPT-5+\timesfm) & 0.783 & 0.725 & 4.0 \\
 & REV(GPT-OSS-120B+\timesfm) & 0.794 & 0.787 & 7.0 \\
 & REV(Gemini-3.1-Flash-Lite+\timesfm) & 0.820 & 2.966 & 19.0 \\
 & REV(Qwen-3-VL-32B+\timesfm) & 0.915 & 1.026 & 15.0 \\
 & REV(Gemini-3-Flash+\timesfm) & 0.950 & 7.349 & 32.5 \\
 & REV(Gemini-2.5-Flash+\timesfm) & 1.151 & 6.716 & 36.0 \\
 & REV(Gemini-3.1-Pro+\timesfm) & 0.949 & 9.375 & 33.0 \\
 & Rev(Gemma-3-4B+\timesfm) & 0.877 & 0.878 & 14.0 \\
\midrule
\multirow{3}{*}{\textit{TSLMs}} & TimeOmni-1-7B & 1.088 & 1.417 & 24.5 \\
 & Aurora & 1.084 & 1.387 & 23.5 \\
 & ChatTS-8B & 1.189 & 1.689 & 27.0 \\
\midrule
\multirow{11}{*}{\textit{LLMs}} & GPT-5 & 0.843 & 1.299 & 16.0 \\
 & Qwen-3-VL-32B & 0.947 & 1.122 & 17.5 \\
 & GPT-OSS-120B & 0.930 & 2.762 & 22.0 \\
 & MiMo-V2-Flash & 0.969 & 3.039 & 27.0 \\
 & Gemini-3.1-Flash-Lite & 0.812 & 3.101 & 20.0 \\
 & Gemma-4-26B-A4B & 0.997 & 4.605 & 31.0 \\
 & Gemma-4-31B & 0.955 & 5.717 & 31.0 \\
 & Gemini-3.1-Pro & 0.917 & 8.890 & 30.5 \\
 & Gemini-3-Flash & 1.022 & 10.014 & 36.0 \\
 & Gemini-2.5-Flash & 1.261 & 13.445 & 41.0 \\
 & Gemma-3-4B & 1.486 & 11.434 & 41.5 \\
\midrule
\multirow{3}{*}{\textit{TSFMs}} & \timesfm-2.5 & 0.788 & 0.729 & 5.0 \\
 & Moirai-2.0 & 0.800 & 0.731 & 7.5 \\
 & Sundial & 1.121 & 3.144 & 30.5 \\
\midrule
\multirow{2}{*}{\textit{Ours}} & \textsc{PostTime}$_{\mathrm{SFT}}$ (Gemma-3-4B) & 0.739 & 0.650 & 2.0 \\
 & \textbf{\method (Gemma-3-4B+\timesfm)} & \textbf{0.738} & \textbf{0.638} & \textbf{1.0} \\
\midrule
Summary & Imp. over \timesfm-2.5 & 6.38\% & 12.43\% & -- \\
\bottomrule
\end{tabular}
}
}
\end{table}

\clearpage
\begin{table}[!p]
\centering
\caption{OOD main results on 11 out-of-domain variables. The SFT-only \method variant is inserted immediately above the final RL-refined \method, which improves over \timesfm-2.5 by 3.93\%nMAE and 11.24\%nMSE, reaches the bestnMSE, and achieves the best average rank. Rows are grouped by the leftmost category column; lower is better, and bold marks the final \method and metric bests.}
\label{tab:main_results_ood}
{\scriptsize
\setlength{\tabcolsep}{2pt}
\renewcommand{\arraystretch}{0.70}
\resizebox{0.98\textwidth}{!}{
\begin{tabular}{@{}llrrr@{}}
\toprule
Group & Method &n MAE $\downarrow$ &n MSE $\downarrow$ & Avg. Rank $\downarrow$ \\
\midrule
\multirow{9}{*}{\textit{Supervised}} & TTS(iTransformer+GPT2) & 1.188 & 1.508 & 35.0 \\
 & TTS(FiLM+GPT2) & 1.637 & 2.186 & 38.0 \\
 & TTS(DLinear+GPT2) & 1.836 & 2.766 & 40.0 \\
 & TTS(Informer+GPT2) & 1.995 & 5.064 & 43.0 \\
 & TTS(Transformer+GPT2) & 1.963 & 3.497 & 41.0 \\
 & TTS(PatchTST+GPT2) & 1.705 & 3.819 & 40.5 \\
 & TTS(FEDformer+GPT2) & 2.342 & 5.923 & 44.0 \\
 & TTS(Autoformer+GPT2) & 2.432 & 7.429 & 45.0 \\
 & TTS(Crossformer+GPT2) & 3.347 & 19.938 & 46.0 \\
\midrule
\multirow{18}{*}{\textit{Training-free}} & ENS(GPT-5+\timesfm) & 0.769 & 0.660 & 4.0 \\
 & ENS(Qwen-3-VL-32B+\timesfm) & 0.901 & 1.083 & 23.0 \\
 & ENS(GPT-OSS-120B+\timesfm) & 0.857 & 0.787 & 16.0 \\
 & ENS(MiMo-V2-Flash+\timesfm) & 0.863 & 0.739 & 14.5 \\
 & ENS(Gemini-3.1-Flash-Lite+\timesfm) & 0.795 & 0.780 & 11.5 \\
 & ENS(Gemma-4-26B-A4B+\timesfm) & 0.877 & 0.879 & 19.0 \\
 & ENS(Gemma-4-31B+\timesfm) & 0.828 & 0.801 & 14.5 \\
 & ENS(Gemini-3.1-Pro+\timesfm) & \textbf{0.742} & 0.637 & 2.0 \\
 & ENS(Gemini-3-Flash+\timesfm) & 0.802 & 0.743 & 11.5 \\
 & ENS(Gemini-2.5-Flash+\timesfm) & 0.856 & 0.852 & 17.0 \\
 & REV(GPT-5+\timesfm) & 0.836 & 0.737 & 12.0 \\
 & REV(GPT-OSS-120B+\timesfm) & 0.950 & 1.331 & 27.5 \\
 & REV(Gemini-3.1-Flash-Lite+\timesfm) & 0.827 & 0.752 & 12.5 \\
 & REV(Qwen-3-VL-32B+\timesfm) & 1.356 & 4.222 & 40.0 \\
 & REV(Gemini-3-Flash+\timesfm) & 0.801 & 0.698 & 9.0 \\
 & REV(Gemini-2.5-Flash+\timesfm) & 0.967 & 1.380 & 28.5 \\
 & REV(Gemini-3.1-Pro+\timesfm) & 0.787 & 0.659 & 6.0 \\
 & Rev(Gemma-3-4B+\timesfm) & 0.894 & 0.883 & 20.0 \\
\midrule
\multirow{3}{*}{\textit{TSLMs}} & TimeOmni-1-7B & 0.777 & 0.673 & 6.0 \\
 & Aurora & 1.025 & 1.113 & 26.5 \\
 & ChatTS-8B & 1.117 & 1.394 & 33.0 \\
\midrule
\multirow{11}{*}{\textit{LLMs}} & GPT-5 & 0.846 & 0.809 & 16.0 \\
 & Qwen-3-VL-32B & 1.163 & 2.395 & 36.5 \\
 & GPT-OSS-120B & 1.027 & 1.177 & 27.5 \\
 & MiMo-V2-Flash & 1.037 & 1.068 & 26.5 \\
 & Gemini-3.1-Flash-Lite & 0.900 & 1.325 & 25.0 \\
 & Gemma-4-26B-A4B & 1.090 & 1.523 & 34.0 \\
 & Gemma-4-31B & 0.999 & 1.239 & 27.5 \\
 & Gemini-3.1-Pro & 0.771 & 0.702 & 6.5 \\
 & Gemini-3-Flash & 0.913 & 0.994 & 22.5 \\
 & Gemini-2.5-Flash & 1.056 & 1.447 & 32.0 \\
 & Gemma-3-4B & 1.306 & 1.944 & 36.5 \\
\midrule
\multirow{3}{*}{\textit{TSFMs}} & \timesfm-2.5 & 0.777 & 0.673 & 6.0 \\
 & Moirai-2.0 & 0.942 & 0.905 & 22.5 \\
 & Sundial & 1.057 & 1.211 & 29.5 \\
\midrule
\multirow{2}{*}{\textit{Ours}} & \textsc{PostTime}$_{\mathrm{SFT}}$ (Gemma-3-4B) & 0.772 & 0.623 & 3.5 \\
 & \textbf{\method (Gemma-3-4B+\timesfm)} & \textbf{0.746} & \textbf{0.597} & \textbf{1.5} \\
\midrule
Summary & Imp. over \timesfm-2.5 & 3.93\% & 11.24\% & -- \\
\bottomrule
\end{tabular}
}
}
\end{table}

\clearpage

%% file: citations.bib
@inproceedings{liu2026rethinking,
  title={Rethinking Multimodal Time-Series Forecasting Evaluation},
  author={Liu, Haoxin and Zhou, Yichen and Sen, Rajat and Prakash, B Aditya and Das, Abhimanyu},
  booktitle={1st ICLR Workshop on Time Series in the Age of Large Models},
  year={2026},
}

@article{guan2025timeomni,
  title={Timeomni-1: Incentivizing complex reasoning with time series in large language models},
  author={Guan, Tong and Meng, Zijie and Li, Dianqi and Wang, Shiyu and Yang, Chao-Han Huck and Wen, Qingsong and Liu, Zuozhu and Siniscalchi, Sabato Marco and Jin, Ming and Pan, Shirui},
  journal={arXiv preprint arXiv:2509.24803},
  year={2025}
}

@article{liu2025can,
  title={How can time series analysis benefit from multiple modalities? a survey and outlook},
  author={Liu, Haoxin and Kamarthi, Harshavardhan and Zhao, Zhiyuan and Xu, Shangqing and Wang, Shiyu and Wen, Qingsong and Hartvigsen, Tom and Wang, Fei and Prakash, B Aditya},
  journal={arXiv preprint arXiv:2503.11835},
  year={2025}
}

@article{das2023decoder,
  title={A decoder-only foundation model for time-series forecasting},
  author={Das, Abhimanyu and Kong, Weihao and Sen, Rajat and Zhou, Yichen},
  journal={arXiv preprint arXiv:2310.10688},
  year={2023}
}

@inproceedings{woo2024unified,
  title={Unified training of universal time series forecasting transformers},
  author={Woo, Gerald and Liu, Chenghao and Kumar, Akshat and Xiong, Caiming and Savarese, Silvio and Sahoo, Doyen},
  booktitle={Forty-first International Conference on Machine Learning},
  year={2024}
}

@inproceedings{merrill2024language,
  title={Language models still struggle to zero-shot reason about time series},
  author={Merrill, Mike A and Tan, Mingtian and Gupta, Vinayak and Hartvigsen, Thomas and Althoff, Tim},
  booktitle={Findings of the Association for Computational Linguistics: EMNLP 2024},
  pages={3512--3533},
  year={2024}
}

@article{tan2024language,
  title={Are language models actually useful for time series forecasting?},
  author={Tan, Mingtian and Merrill, Mike A and Gupta, Vinayak and Althoff, Tim and Hartvigsen, Thomas},
  journal={Advances in Neural Information Processing Systems},
  volume={37},
  pages={60162--60191},
  year={2024}
}

@inproceedings{liu2025sundial,
  title={Sundial: A Family of Highly Capable Time Series Foundation Models},
  author={Liu, Yong and Qin, Guo and Shi, Zhiyuan and Chen, Zhi and Yang, Caiyin and Huang, Xiangdong and Wang, Jianmin and Long, Mingsheng},
  booktitle={International Conference on Machine Learning},
  pages={39295--39317},
  year={2025},
  organization={PMLR}
}

@article{wu2025aurora,
  title={Aurora: Towards universal generative multimodal time series forecasting},
  author={Wu, Xingjian and Jin, Jianxin and Qiu, Wanghui and Chen, Peng and Shu, Yang and Yang, Bin and Guo, Chenjuan},
  journal={arXiv preprint arXiv:2509.22295},
  year={2025}
}

@article{xie2025chatts,
  title={ChatTS: Aligning Time Series with LLMs via Synthetic Data for Enhanced Understanding and Reasoning},
  author={Xie, Zhe and Li, Zeyan and He, Xiao and Xu, Longlong and Wen, Xidao and Zhang, Tieying and Chen, Jianjun and Shi, Rui and Pei, Dan},
  journal={Proceedings of the VLDB Endowment},
  volume={18},
  number={8},
  pages={2385--2398},
  year={2025},
  publisher={VLDB Endowment}
}

@article{kong2025timemqa,
  title={Time-MQA: Time Series Multi-Task Question Answering with Context Enhancement},
  author={Kong, Yaxuan and Yang, Yiyuan and Hwang, Yoontae and Du, Wenjie and Zohren, Stefan and Wang, Zhangyang and Jin, Ming and Wen, Qingsong},
  journal={arXiv preprint arXiv:2503.01875},
  year={2025},
  note={ACL 2025 (Main)}
}

@misc{liu2025timer1,
      title={Time Series Forecasting as Reasoning: A Slow-Thinking Approach with Reinforced LLMs}, 
      author={Yitong Zhou and Yucong Luo and Mingyue Cheng and Qi Liu and Jiahao Wang and Daoyu Wang and Enhong Chen},
      year={2026},
      eprint={2506.10630},
      archivePrefix={arXiv},
      primaryClass={cs.LG},
      url={https://arxiv.org/abs/2506.10630}, 
}

@article{ansari2025chronos,
  title={Chronos-2: From univariate to universal forecasting},
  author={Ansari, Abdul Fatir and Shchur, Oleksandr and K{\"u}ken, Jaris and Auer, Andreas and Han, Boran and Mercado, Pedro and Rangapuram, Syama Sundar and Shen, Huibin and Stella, Lorenzo and Zhang, Xiyuan and others},
  journal={arXiv preprint arXiv:2510.15821},
  year={2025}
}

@article{Wei2021,
  author  = {Wei, Jason and Bosma, Maarten and Vincent, Y. and Wu, Kelvin and Chang, Ming-Wei and Salvucci, Andrew and Lou, Daniel and Vasudevan, Vijay and Ni, Tianqi and Hou, Le and Dai, Andrew M. and Le, Quoc V.},
  title   = {Finetuned Language Models are Zero-Shot Learners},
  journal = {arXiv preprint arXiv:2109.01652},
  year    = {2021},
  url     = {https://arxiv.org/abs/2109.01652}
}

@inproceedings{
li2026language,
title={Language in the Flow of Time: Time-Series-Paired Texts Weaved into a Unified Temporal Narrative},
author={Zihao Li and Xiao Lin and Zhining Liu and Jiaru Zou and Ziwei Wu and Lecheng Zheng and Dongqi Fu and Yada Zhu and Hendrik Hamann and Hanghang Tong and Jingrui He},
booktitle={The Fourteenth International Conference on Learning Representations},
year={2026},
url={https://openreview.net/forum?id=a1zBg9cBvt}
}

@article{ouyang2022training,
  title={Training language models to follow instructions with human feedback},
  author={Ouyang, Long and Wu, Jeffrey and Jiang, Xu and Almeida, Diogo and Wainwright, Carroll and Mishkin, Pamela and Zhang, Chong and Agarwal, Sandhini and Slama, Katarina and Ray, Alex and others},
  journal={Advances in neural information processing systems},
  volume={35},
  pages={27730--27744},
  year={2022}
}

@article{chung2024scaling,
  title={Scaling instruction-finetuned language models},
  author={Chung, Hyung Won and Hou, Le and Longpre, Shayne and Zoph, Barret and Tay, Yi and Fedus, William and Li, Yunxuan and Wang, Xuezhi and Dehghani, Mostafa and Brahma, Siddhartha and others},
  journal={Journal of Machine Learning Research},
  volume={25},
  number={70},
  pages={1--53},
  year={2024}
}

@article{wei2022chain,
  title={Chain-of-thought prompting elicits reasoning in large language models},
  author={Wei, Jason and Wang, Xuezhi and Schuurmans, Dale and Bosma, Maarten and Xia, Fei and Chi, Ed and Le, Quoc V and Zhou, Denny and others},
  journal={Advances in neural information processing systems},
  volume={35},
  pages={24824--24837},
  year={2022}
}

@article{zelikman2022star,
  title={Star: Bootstrapping reasoning with reasoning},
  author={Zelikman, Eric and Wu, Yuhuai and Mu, Jesse and Goodman, Noah},
  journal={Advances in Neural Information Processing Systems},
  volume={35},
  pages={15476--15488},
  year={2022}
}

@article{guo2025deepseek,
  title={Deepseek-r1: Incentivizing reasoning capability in llms via reinforcement learning},
  author={Guo, Daya and Yang, Dejian and Zhang, Haowei and Song, Junxiao and Wang, Peiyi and Zhu, Qihao and Xu, Runxin and Zhang, Ruoyu and Ma, Shirong and Bi, Xiao and others},
  journal={arXiv preprint arXiv:2501.12948},
  year={2025}
}

@article{shao2024deepseekmath,
  title={Deepseekmath: Pushing the limits of mathematical reasoning in open language models},
  author={Shao, Zhihong and Wang, Peiyi and Zhu, Qihao and Xu, Runxin and Song, Junxiao and Bi, Xiao and Zhang, Haowei and Zhang, Mingchuan and Li, YK and Wu, Yang and others},
  journal={arXiv preprint arXiv:2402.03300},
  year={2024}
}

@article{liu2024time,
  title={Time-mmd: Multi-domain multimodal dataset for time series analysis},
  author={Liu, Haoxin and Xu, Shangqing and Zhao, Zhiyuan and Kong, Lingkai and Kamarthi, Harshavardhan and Sasanur, Aditya B and Sharma, Megha and Cui, Jiaming and Wen, Qingsong and Zhang, Chao and others},
  journal={Advances in Neural Information Processing Systems},
  volume={37},
  pages={77888--77933},
  year={2024}
}

@inproceedings{
williams2025context,
title={Context is Key: A Benchmark for Forecasting with Essential Textual Information},
author={Andrew Robert Williams and Arjun Ashok and {\'E}tienne Marcotte and Valentina Zantedeschi and Jithendaraa Subramanian and Roland Riachi and James Requeima and Alexandre Lacoste and Irina Rish and Nicolas Chapados and Alexandre Drouin},
booktitle={Forty-second International Conference on Machine Learning},
year={2025},
url={https://openreview.net/forum?id=ih2WuBT1Fn}
}

@article{kamarthi2024large,
  title={Large pre-trained time series models for cross-domain time series analysis tasks},
  author={Kamarthi, Harshavardhan and Prakash, B Aditya},
  journal={Advances in Neural Information Processing Systems},
  volume={37},
  pages={56190--56214},
  year={2024}
}

@inproceedings{liu2024lingkai,
  title={Lingkai Kong, Zhiyuan Zhao, Chao Zhang, and B Aditya Prakash. Time-series forecasting for out-of-distribution generalization using invariant learning},
  author={Liu, Haoxin and Kamarthi, Harshavardhan},
  booktitle={Proceedings of the 41st International Conference on Machine Learning},
  pages={31312--31325},
  year={2024}
}

@inproceedings{liu-etal-2025-picture,
    title = "A Picture is Worth A Thousand Numbers: Enabling {LLM}s Reason about Time Series via Visualization",
    author = "Liu, Haoxin  and
      Liu, Chenghao  and
      Prakash, B. Aditya",
    editor = "Chiruzzo, Luis  and
      Ritter, Alan  and
      Wang, Lu",
    booktitle = "Proceedings of the 2025 Conference of the Nations of the Americas Chapter of the Association for Computational Linguistics: Human Language Technologies (Volume 1: Long Papers)",
    month = apr,
    year = "2025",
    address = "Albuquerque, New Mexico",
    publisher = "Association for Computational Linguistics",
    url = "https://aclanthology.org/2025.naacl-long.383/",
    doi = "10.18653/v1/2025.naacl-long.383",
    pages = "7486--7518",
    ISBN = "979-8-89176-189-6"
}

@article{liu2601futurex,
  title={FutureX-Pro: Extending Future Prediction to High-Value Vertical Domains, 2026},
  author={Liu, Jiashuo and Chen, Siyuan and Wang, Zaiyuan and Zeng, Zhiyuan and Guo, Jiacheng and Hu, Liang and Yin, Lingyue and Huang, Suozhi and Hao, Wenxin and Yang, Yang and others},
  year = "2026",
  journal={URL https://arxiv. org/abs/2601.12259}
}
